%% file: main.tex
\documentclass{article}

\usepackage{PRIMEarxiv}

\usepackage[utf8]{inputenc} 
\usepackage[T1]{fontenc}    
\usepackage{url}            
\usepackage{booktabs}       
\usepackage{amsmath,amssymb,amsfonts}
\usepackage{nicefrac}       
\usepackage{microtype}      
\usepackage{fancyhdr}       
\usepackage{graphicx}       
\usepackage[dvipsnames]{xcolor}
\usepackage{xspace}
\usepackage[numbers,sort&compress]{natbib}
\usepackage[inline]{enumitem}
\usepackage{caption}
\usepackage{multirow}
\usepackage{placeins}
\graphicspath{{imgs/}}     
\definecolor{cvprblue}{rgb}{0.21,0.49,0.74}
\usepackage[pagebackref,breaklinks,colorlinks,allcolors=cvprblue]{hyperref}
\usepackage[capitalize]{cleveref}
\crefname{section}{Sec.}{Secs.}
\Crefname{section}{Section}{Sections}
\crefname{table}{Tab.}{Tabs.}
\Crefname{table}{Table}{Tables}
\crefname{figure}{Fig.}{Figs.}
\crefname{equation}{Eq.}{Eqs.}

\newcommand{\ours}{EchoTrace\xspace}
\newcommand{\dsone}{Diffusion-10\xspace}
\newcommand{\dstwo}{Mixed-10\xspace}
\newcommand{\Khat}{\hat{K}}
\newcommand{\frozenicon}{\raisebox{-0.3ex}{\includegraphics[height=1.9ex]{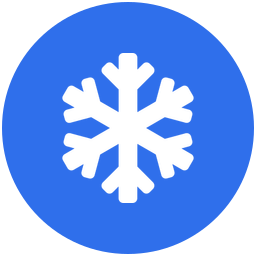}}}
\newcommand{\trainedicon}{\raisebox{-0.3ex}{\includegraphics[height=1.9ex]{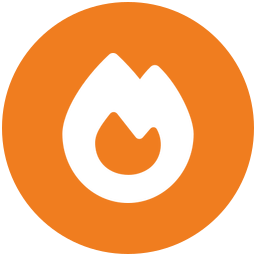}}}
\newcommand{\eg}{\emph{e.g.}\xspace}
\newcommand{\ie}{\emph{i.e.}\xspace}
\newcommand{\etal}{\emph{et al.}\xspace}

\title{EchoTrace: Discovering Unknown Face Generators\\Through Consistent Latent Traces}

\author{
    \href{https://orcid.org/0009-0001-1284-6631}{\includegraphics[scale=0.06]{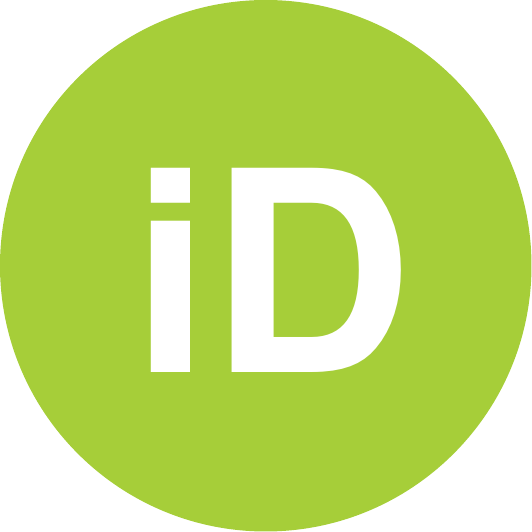}\hspace{1mm}Claudio Giusti} \\
  Department of Mathematics and Computer Science\\
  University of Catania\\
  Catania, CT 95125 \\
  \texttt{claudio.giusti@studium.unict.it} \\
   \And
  \href{https://orcid.org/0000-0001-8315-351X}{\includegraphics[scale=0.06]{orcid.pdf}\hspace{1mm}Luca Guarnera} \\
  Department of Mathematics and Computer Science\\
  University of Catania\\
  Catania, CT 95125 \\
  \texttt{luca.guarnera@unict.it} \\
  \And
  \href{https://orcid.org/0000-0001-6127-2470}{\includegraphics[scale=0.06]{orcid.pdf}\hspace{1mm}Sebastiano Battiato} \\
  Department of Mathematics and Computer Science\\
  University of Catania\\
  Catania, CT 95125 \\
  \texttt{sebastiano.battiato@unict.it} \\
}

\begin{document}
\maketitle

\input{sec/0_abstract}


\begin{figure}[!htb]
  \centering
  \includegraphics[width=\textwidth]{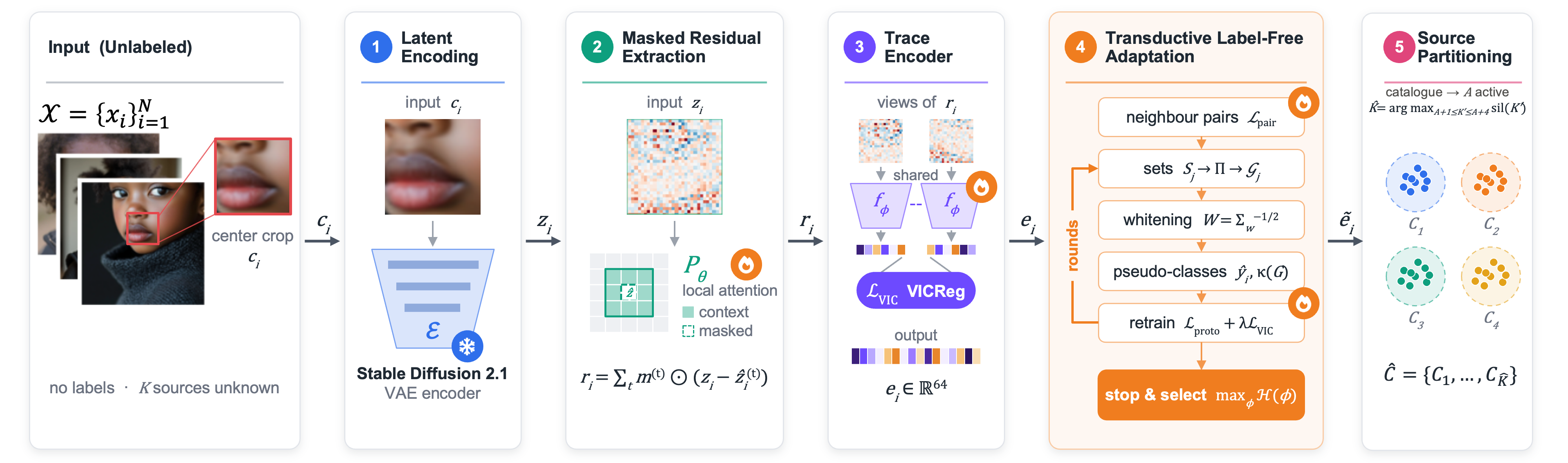}
  \caption{\textbf{Overview of \ours.}
  \textbf{Latent Encoding} (\cref{sec:latent}): a frozen autoencoder $\mathcal{E}$ maps the crop
  $c_i$ to $z_i$.
  \textbf{Masked Residual Extraction} (\cref{sec:residual}): $P_\theta$ predicts each masked
  position; the residual $r_i$ keeps the generator trace.
  \textbf{Trace Encoder} (\cref{sec:encoder}): $f_\phi$ embeds $r_i$ into $e_i$.
  \textbf{Transductive Label-Free Adaptation} (\cref{sec:adaptation}): mutual neighbors, then
  recurring cohesive groups as pseudo-classes, retrain $f_\phi$; $\mathcal{H}$ selects the model.
  \textbf{Source Partitioning} (\cref{sec:partition}): a catalog of identities gives $A$, and the
  silhouette chooses $\Khat$ and $\hat{\mathcal{C}}$ above it. \frozenicon~frozen, never updated; \trainedicon~trained, without labels.}
  \label{fig:pipeline}
\end{figure}

\input{sec/1_intro}
\input{sec/2_related}
\input{sec/3_method}
\input{sec/4_experiments}
\input{sec/5_conclusion}

\FloatBarrier
\bibliographystyle{unsrtnat}
\bibliography{main}

\end{document}

%% file: sec/0_abstract.tex
\begin{abstract}
Forensic analysis of AI-generated images increasingly requires knowing which generator produced
an image, not only whether it is synthetic. In practice, collections of synthetic images come with
no labels, may include unseen generators, and contain an unknown number of sources. We address this
\emph{label-free open-world source attribution} problem: partitioning an unlabeled set of
generated images by source generator, with the number of generators unknown. We focus on synthetic
faces, the main vector of fake profiles and impersonation, where generators must be told apart by
their trace rather than by what they depict. We propose \ours,
which isolates generator traces as the residuals of a masked local predictor in the latent space of
a frozen autoencoder. It embeds them with a self-supervised encoder, adapts the representation transductively to the
unlabeled collection through mutual neighbors and label-free pseudo-classes, and estimates the
number of sources from a catalog of identities that recur in the collection. No source label is
used for training, model selection, or stopping. On two datasets of ten face generators each,
\ours roughly doubles the adjusted Rand index of the strongest label-free method and estimates the
number of generators to within about one on average, also when all generators share the same resolution, where prior
label-free methods collapse to a few clusters. Code will be released upon acceptance.
\end{abstract}

%% file: sec/1_intro.tex
\section{Introduction}
\label{sec:intro}

Telling real from synthetic is no longer enough. Photorealistic faces are produced by dozens of
generators, from open latent diffusion models~\cite{rombach2022ldm} to commercial services, and a
forensic analyst must establish which generator produced an image to link images to a
common origin, trace a campaign, or hold a service accountable. Yet most work targets
detection~\cite{wang2020cnn,ojha2023universal,ricker2024aeroblade,park2025community}, and
attribution is treated as closed-set classification over known, labeled
generators~\cite{marra2019gans,yu2019attributing,sha2023defake,protoleaknet}. Real cases break
both assumptions: a seized collection carries no labels, contains generators never seen before, and
the number of sources is itself what the analyst needs to find out.

We tackle this setting directly: partition an unlabeled collection of AI-generated images by
source generator, \emph{with no labels and no knowledge of how many generators are present}.
Existing methods fall short of it. Open-set methods~\cite{girish2021discovery,yang2023pose} reject
unknown generators but leave them unorganized. Open-world methods~\cite{cpl,cdal,cal} discover new
generators only when given labeled known ones. The few label-free approaches either collapse to two
or three clusters~\cite{fsd} or rely on foundation features dominated by content~\cite{drowcula}.
Content is useless here: all images are portraits, often from the same prompts, and several are closely
related services (\eg, Flux.1 and Flux.1.1~Pro). Only the generator's own trace can separate them.
Restricting the problem to one semantic class makes it well posed, since generators then differ by
their trace rather than by what they depict. The setting is also naturally transductive: an investigation analyzes a seized
collection or a campaign as a whole, which provides the pool to adapt to: its generators are unknown, \ie unlabeled
and unnamed, but present in it.

Our key insight is that the task needs two distinct mechanisms: a trace that survives
post-processing, and a representation in which that trace, not content, drives similarity.
\ours (\cref{fig:pipeline}) obtains the first from \emph{what a model of local image structure
fails to predict}. It encodes each image with a frozen latent autoencoder~\cite{rombach2022ldm},
predicts every latent position from its masked neighborhood, and keeps the residual, the
counterpart of a sensor fingerprint~\cite{protoleaknet,lukas2006prnu}. Compared with raw latents,
residuals give a higher clean accuracy and keep about twice as much under heavy post-processing (\cref{sec:ablation}). A self-supervised trace
encoder~\cite{bardes2022vicreg} embeds the residuals. For the second mechanism, a transductive
adaptation turns the unlabeled collection into its own supervision. Mutual nearest neighbors first
sharpen the local structure; then groups that recur across many sampled image sets become
pseudo-classes, the within-group scatter is whitened away, and the encoder is retrained, with
stopping and model selection driven by a label-free cohesion criterion. The number of sources is
finally anchored to a catalog of identities that recur in the collection, and refined by the
silhouette~\cite{rousseeuw1987silhouettes}. No source label is ever used for training, model
selection or stopping.

On two face datasets of ten generators each, \ours roughly \emph{doubles} the adjusted Rand index of
the best existing label-free method and estimates the number of generators to within about one on average, also
when all generators share the same native resolution, while prior label-free approaches collapse to
a handful of clusters. It also holds its own against open-world methods trained on ten \emph{labeled}
generators: it outperforms them on one dataset, and on the other is surpassed only by a method that
is also given the true number of generators.

\noindent Our contributions are:
\begin{itemize}[leftmargin=*,itemsep=0pt,topsep=2pt]
  \item We introduce \emph{label-free open-world source attribution}, together with an evaluation
  protocol on sampled image sets of unknown composition and partition, per-source and
  representation metrics.
  \item We propose \ours, which isolates generator traces as masked-prediction residuals in a
  frozen latent space, adapts their representation with mutual neighbors and label-free
  pseudo-classes, and estimates $K$ from a catalog of recurring identities.
  \item We show that \ours yields the best partitions among label-free methods, the most accurate
  estimate of $K$ across set compositions (\cref{fig:kgen}), and a representation that separates generators
  better than semi-supervised methods trained on labeled generators. It is also the most robust
  label-free method under chained post-processing.
\end{itemize}

\noindent \cref{sec:related} reviews related work, \cref{sec:method} formalizes the problem and
describes \ours, \cref{sec:experiments} evaluates it, \cref{sec:discussion} discusses its
limitations, and \cref{sec:conclusion} concludes.

%% file: sec/2_related.tex
\section{Related Work}
\label{sec:related}

Source attribution has moved from supervised classification over known generators towards
settings with fewer assumptions; we follow this progression up to the label-free methods closest to
ours.

\paragraph{Detection and closed-set attribution.}
Synthetic-image detectors exploit generator artefacts~\cite{wang2020cnn,frank2020frequency,
corvi2023detection}, pretrained features~\cite{ojha2023universal,cozzolino2024raising},
reconstruction errors~\cite{wang2023dire,ricker2024aeroblade} or many training
generators~\cite{park2025community}, but give a binary answer. Attribution builds on the fact that
generators leave fingerprints, as cameras do through sensor noise~\cite{lukas2006prnu}. These
fingerprints can be extracted from residuals~\cite{marra2019gans}, learned~\cite{yu2019attributing},
or defined on the data manifold~\cite{song2024manifpt,song2025riemannian}. They support model
parsing~\cite{asnani2023reverse}, the attribution of deepfake and text-to-image
models~\cite{guarnera2022exploitation,sha2023defake}, and attribution from latent signal-leak
traces~\cite{protoleaknet,everaert2024signalleak}. All are supervised over a closed set of labeled
generators.

\paragraph{Open-set and open-world attribution.}
Open-set methods reject unseen generators~\cite{yang2023pose,wang2023origin}, and
Girish~\etal~\cite{girish2021discovery} cluster the rejected samples to discover new GANs.
Open-world deepfake attribution~\cite{cpl} adopts the category-discovery
setting~\cite{han2019dtc,vaze2022gcd}: a labeled set of known forgery types is given together with
unlabeled images of known and novel types. CPL~\cite{cpl} learns from both with contrastive
pseudo-labels weighted by confidence.
CDAL~\cite{cdal} builds on it and decouples the attention to model-specific traces from the
attention to source content through counterfactual interventions. It classifies over a fixed head
whose size must equal the total number of known and novel classes, so $K$ must be given.
CAL~\cite{cal} treats known and novel classes asymmetrically, calibrating pseudo-labels by
confidence so that novel classes are not absorbed by known ones, and estimates the number of novel
classes by pruning redundant prototypes. These methods require labeled generators, and their
accuracy depends on how known and unknown generators relate. We compare against CDAL and CAL, the
most recent of them; CAL is the only one that estimates $K$.

\paragraph{Label-free grouping.}
Forensic self-descriptions (FSD)~\cite{fsd} learn, from real images only, predictive filters whose
residuals describe the forensic microstructure of each image; the resulting descriptor supports
zero-shot detection, open-set attribution and clustering by source, but it is fixed once learned
and does not adapt to the collection being analyzed. DROWCULA~\cite{drowcula} embeds images with
DINOv2~\cite{oquab2024dinov2}, reduces the embeddings with UMAP and clusters them with $k$-means,
choosing $K$ by the silhouette. Deep clustering~\cite{caron2018deepcluster,vangansbeke2020scan,
adaloglou2023temi} and foundation features~\cite{oquab2024dinov2,radford2021clip} are trained to
capture semantic content, which does not separate generators rendering the same content. The
closest forensic analog is blind camera clustering~\cite{bloy2008blind}, which groups photographs
by sensor noise without knowing the cameras.

\paragraph{Positioning of \ours.}
Unlike these methods, \ours uses no source label at any stage, including model selection and
stopping, and estimates $K$ from the data. Its trace comes from native-resolution crops in a frozen
latent space, and its representation is adapted to the unlabeled collection itself. We compare
with the most recent methods for this task whose code is publicly available; later works either
address different settings, such as detection or closed-set attribution, or do not release the code
needed for a faithful comparison.

%% file: sec/3_method.tex
\section{Method}
\label{sec:method}

\ours is a framework of five stages, shown in \cref{fig:pipeline}. After formalizing the problem,
we describe each stage in turn, following an image from its pixels to the estimated partition.

\paragraph{Problem statement.}
Let $\mathcal{X}=\{x_i\}_{i=1}^{N}$ be an unlabeled collection of synthetic images, each produced
by an unknown generator $y_i\in\{1,\dots,K\}$, where the number of generators $K$ is also unknown.
The goal is a partition
\begin{equation}
  \begin{split}
    \hat{\mathcal{C}}&=\{C_1,\dots,C_{\Khat}\},\\
    \textstyle\bigcup_{k} C_k&=\mathcal{X},\qquad C_k\cap C_{k'}=\emptyset\ \ (k\neq k'),
  \end{split}
  \label{eq:partition}
\end{equation}
where each cluster $C_k\subseteq\mathcal{X}$ is a group of images attributed to the same source,
$\Khat$ is the number of clusters, estimated from the data, and the two conditions require every
image to belong to exactly one cluster. Clusters carry no generator name: attribution here means
grouping images by source, not naming it. The ideal partition has one cluster per generator,
$C_k=\{x_i:y_i=k\}$ up to a permutation of the indices, and therefore $\Khat=K$. The labels $y_i$
are never observed by \ours and are used only to evaluate $\hat{\mathcal{C}}$.

\subsection{Latent Encoding}
\label{sec:latent}
Each image is center-cropped at its native resolution, without resampling, which would blur the
high-frequency traces we are after: $c_i=\mathrm{crop}_{s}(x_i)\in[-1,1]^{3\times s\times s}$, with
$s=256$ and intensities rescaled to $[-1,1]$. The crop is mapped to the latent space of the frozen
autoencoder $\mathcal{E}$ of Stable Diffusion~2.1~\cite{rombach2022ldm}:
\begin{equation}
  z_i=\big(\mathcal{E}(c_i)-\mu_z\big)\oslash\sigma_z\in\mathbb{R}^{4\times H_z\times W_z},
  \label{eq:latent}
\end{equation}
with spatial size $H_z=W_z=s/8=32$, where $\mu_z,\sigma_z\in\mathbb{R}^{4}$ are per-channel training statistics, broadcast over
positions, and $\oslash$ is element-wise division. $\mathcal{E}$ is never updated. Its latent space compresses the image content while keeping the
statistics that latent generators shape directly.

\subsection{Masked Residual Extraction}
\label{sec:residual}
A small transformer $P_\theta$ with parameters $\theta$ predicts hidden positions of the latent of
\cref{eq:latent} from their visible neighborhood. Given a binary mask $m\in\{0,1\}^{H_z\times W_z}$ ($m_{uv}=1$ marks a hidden
position), the prediction is
\begin{equation}
  \hat{z}_i=P_\theta\big(z_i\odot(1-m),\,m\big),
  \label{eq:predict}
\end{equation}
where $\odot$ broadcasts $m$ over the four channels and hidden positions are replaced by a learned
mask token. $P_\theta$ uses two blocks of local self-attention restricted to $5\times5$ windows, so
its prediction captures only local structure, the content-driven part of the signal. Training
samples masks of $4\times4$ blocks covering a fraction $\rho=0.25$ of the positions and minimizes
the error on hidden positions only:
\begin{equation}
  \mathcal{L}_{\mathrm{pred}}(\theta)=\frac{\big\|\,m\odot(\hat{z}_i-z_i)\,\big\|_2^2}
  {4\,\|m\|_1}.
  \label{eq:predloss}
\end{equation}
At extraction time we use a fixed set of $M$ covering masks $\{m^{(t)}\}_{t=1}^{M}$ that hide every
position exactly once ($\sum_t m^{(t)}=\mathbf{1}$), so that each position is predicted by \cref{eq:predict} without
seeing itself. Denoting by $\hat{z}_i^{(t)}$ the prediction of \cref{eq:predict} with mask
$m^{(t)}$, the residual map is
\begin{equation}
  r_i=\sum_{t=1}^{M} m^{(t)}\odot\big(z_i-\hat{z}_i^{(t)}\big)\in\mathbb{R}^{4\times H_z\times W_z}.
  \label{eq:residual}
\end{equation}
$r_i$ keeps what local context cannot explain, the counterpart of a sensor
fingerprint~\cite{lukas2006prnu}. Compared with the raw latent $z_i$, it yields a higher clean accuracy and a trace that
degrades far less under post-processing (\cref{sec:ablation}).
Invariance to content is instead obtained by the adaptation of \cref{sec:adaptation}.

\subsection{Trace Encoder}
\label{sec:encoder}
A fully convolutional trace encoder $f_\phi$ (three $3\times3$ convolutions with GELU and global
average pooling) maps a residual map (\cref{eq:residual}), or any crop of it, to $h=f_\phi(r)\in\mathbb{R}^{d}$,
$d=64$.
It is trained with VICReg~\cite{bardes2022vicreg} through a projector $g_\psi$. For a batch of $B$
residuals, two random $16\times16$ crops $r^{\alpha}_i, r^{\beta}_i$ give projections
$u_i=g_\psi(f_\phi(r^{\alpha}_i))$ and $u'_i=g_\psi(f_\phi(r^{\beta}_i))$, stacked in
$U,U'\in\mathbb{R}^{B\times d}$:
\begin{equation}
  \begin{split}
    \mathcal{L}_{\mathrm{VIC}}={}&\lambda_{\mathrm{inv}}\,\tfrac{1}{B}\textstyle\sum_i\|u_i-u'_i\|_2^2\\
    &+\lambda_{\mathrm{var}}\big[\ell_{\mathrm{var}}(U)+\ell_{\mathrm{var}}(U')\big]\\
    &+\lambda_{\mathrm{cov}}\big[\ell_{\mathrm{cov}}(U)+\ell_{\mathrm{cov}}(U')\big],
  \end{split}
  \label{eq:vicreg}
\end{equation}
where $\ell_{\mathrm{var}}(U)=\frac{1}{d}\sum_{l}\max\!\big(0,1-\sqrt{\mathrm{Var}(U_{:,l})+\epsilon}\big)$ keeps every
dimension active, $\ell_{\mathrm{cov}}(U)=\frac{1}{d}\sum_{l\neq l'}[\mathrm{Cov}(U)]_{ll'}^2$ decorrelates
dimensions, $U_{:,l}$ is the $l$-th column of $U$, $\mathrm{Var}(\cdot)$ is its variance over the
batch, $\mathrm{Cov}(U)\in\mathbb{R}^{d\times d}$ is the covariance matrix of the columns of $U$,
$\epsilon$ is a small constant for numerical stability, and $(\lambda_{\mathrm{inv}},\lambda_{\mathrm{var}},\lambda_{\mathrm{cov}})=(25,25,1)$.
The embedding of image $i$ averages $f_\phi$ over the nine $16\times16$
crops $r_i^{(q)}$ at offsets $\{0,8,16\}^2$, standardizes it with training mean and
standard deviation $(\mu_h,\sigma_h)$ and normalizes it:
\begin{equation}
  \begin{split}
    \bar{h}_i&=\Big(\tfrac{1}{9}\textstyle\sum_{q=1}^{9}f_\phi\big(r_i^{(q)}\big)-\mu_h\Big)\oslash\sigma_h,\\
    e_i&=\bar{h}_i/\|\bar{h}_i\|_2 .
  \end{split}
  \label{eq:embedding}
\end{equation}

\subsection{Transductive Label-Free Adaptation}
\label{sec:adaptation}
The space of \cref{eq:embedding} carries source information but is still dominated by
content-related variation. We adapt it to the unlabeled collection in two stages: a neighbor stage
that sharpens the local structure, and a self-training loop of up to $R$ rounds that organizes it
into groups.

\paragraph{Mutual-neighbor pre-training.}
Generators are recognizable locally before they form groups: the nearest neighbor of an image
usually comes from the same generator. We first strengthen this local structure. Mutual nearest
neighbors in the unlabeled pool give the positive pairs
$\mathcal{Q}=\{(i,j):j\in\mathcal{N}_{k_q}(i),\ i\in\mathcal{N}_{k_q}(j)\}$, where
$\mathcal{N}_{k_q}(i)$ are the $k_q$ nearest neighbors of $e_i$ by cosine similarity. For a batch
of $B$ pairs $(i_p,j_p)$, let $o_p$ and $o'_p$ be the normalized outputs of $f_\phi$ on a random
crop of $r_{i_p}$ and of $r_{j_p}$. The encoder minimizes
\begin{equation}
  \begin{split}
    \mathcal{L}_{\mathrm{pair}}={}&-\tfrac{1}{2B}\textstyle\sum_{p}\Big[\log\sigma_p\big(O\,o'_p/T\big)
    +\log\sigma_p\big(O'\,o_p/T\big)\Big]\\
    &+\lambda\,\mathcal{L}_{\mathrm{VIC}},
  \end{split}
  \label{eq:pair}
\end{equation}
where $O,O'\in\mathbb{R}^{B\times d}$ stack the $o_p$ and the $o'_p$, $\sigma_p(\cdot)$ is the
$p$-th entry of the softmax, $T$ a temperature and $\lambda$ weights the VICReg term of
\cref{eq:vicreg}, computed on two crops of the same image. The pairs are recomputed from the
current embeddings at every round of this stage; its encoder initializes the loop below.

\paragraph{Image sets and provisional partitions.}
Each round samples $J$ image sets $S_j\subset\mathcal{X}$ of $n$ images. To give all sets a shared
vocabulary, we compute $N_a$ global anchors as the medoids of a $k$-means on $\{e_i\}$, and fill
each set with the nearest neighbors of a random subset of 2 to 6 anchors. Sampling uses only the
embeddings. Each set is split by a provisional partitioner $\Pi$, a mutual $k_{\mathrm{nn}}$-NN graph ($k_{\mathrm{nn}}=15$)
partitioned by greedy modularity optimization with cluster merging:
$\mathcal{G}_j=\Pi(S_j)=\{G_{j,1},G_{j,2},\dots\}$. $\Pi$ tends to under-segment, and a group merging two generators would teach the encoder that they are one source; sampling around few anchors limits this.

\paragraph{Within-group whitening.}
Variation \emph{inside} a provisional group does not change the source and is mostly content. We
estimate its covariance with the Ledoit--Wolf shrinkage estimator $\mathrm{LW}(\cdot)$~\cite{ledoit2004wolf},
\begin{equation}
  \begin{split}
    \Sigma_w&=\mathrm{LW}\big(\{\,e_i-\bar{e}_{G}:i\in G,\ G\in\mathcal{G}_j,\ j\leq J_w\,\}\big),\\
    W&=\Sigma_w^{-1/2},
  \end{split}
  \label{eq:whitening}
\end{equation}
where $\bar{e}_G$ is the mean embedding of group $G$, $J_w$ the number of sets used, and we work in the whitened space
$\tilde{e}_i=(W e_i)/\|W e_i\|_2$, which suppresses the dominant within-source directions.

\paragraph{Pseudo-classes.}
For a group $G$ we define its prototype $\mu_G$ and its cohesion $\kappa(G)$ as
\begin{equation}
  \mu_G=\frac{\sum_{i\in G}\tilde{e}_i}{\big\|\sum_{i\in G}\tilde{e}_i\big\|_2},\quad
  \kappa(G)=\frac{1}{|G|}\sum_{i\in G}\tilde{e}_i^{\top}\mu_G .
  \label{eq:cohesion}
\end{equation}
A group becomes a candidate when it contains at least $n_{\min}$ images, $|G|\geq n_{\min}$, and its cohesion is above the lowest quartile
of the round. A prototype catalog, updated over all sets of the round with similarity threshold
$\tau$, links candidates from different sets that describe the same region. A candidate is
accepted if its prototype matches a catalog entry with cosine similarity above $\tau+\delta$. Its
images then receive the pseudo-label $\hat{y}_i\in\{1,\dots,K_{\mathrm{ps}}\}$ of that entry, $K_{\mathrm{ps}}$ being the number of
catalog entries, so
that pseudo-classes are unified \emph{across} sets rather than defined within each one.

\paragraph{Retraining.}
The encoder is retrained on the set $\mathcal{P}$ of pseudo-labeled images with a prototypical loss
over learnable prototypes $\{p_k\}_{k=1}^{K_{\mathrm{ps}}}$, initialized at the class means, and
with VICReg as a regularizer:
\begin{equation}
  \mathcal{L}_{\mathrm{proto}}=-\frac{1}{|\mathcal{P}|}\sum_{i\in\mathcal{P}}
  \log\frac{\exp\!\big(\omega_{i,\hat{y}_i}/T\big)}{\sum_{k}\exp\!\big(\omega_{i,k}/T\big)},
  \label{eq:proto}
\end{equation}
\begin{equation}
  \mathcal{L}_{\mathrm{adapt}}=\mathcal{L}_{\mathrm{proto}}+\lambda\,\mathcal{L}_{\mathrm{VIC}},
  \label{eq:adapt}
\end{equation}
where $\omega_{i,k}=f_\phi(r_i)^{\top}p_k/\big(\|f_\phi(r_i)\|_2\,\|p_k\|_2\big)$ is the cosine similarity
between image $i$ and prototype $k$, $T$ is a temperature, $\lambda$ weights the VICReg term, and \cref{eq:proto} is
applied to both views of each image. After retraining, embeddings are recomputed
(\cref{eq:embedding}), $W$ is re-estimated (\cref{eq:whitening}), and the next round starts.

\paragraph{Label-free stopping and model selection.}
Without labels, neither validation accuracy nor clustering quality is available. We select the
encoder with a cohesion criterion computed on $J_v$ validation sets $S^{v}_j$ sampled uniformly at
random:
\begin{equation}
  \begin{split}
    \mathcal{H}(\phi)&=\frac{\sum_{j=1}^{J_v}|\mathcal{G}^{v}_j|\,\bar{\kappa}_j}{\sum_{j=1}^{J_v}|\mathcal{G}^{v}_j|},\\
    \bar{\kappa}_j&=\frac{\sum_{G\in\mathcal{G}^{v}_j}|G|\,\kappa(G)}{\sum_{G\in\mathcal{G}^{v}_j}|G|},
  \end{split}
  \label{eq:criterion}
\end{equation}
where $\kappa(G)$ is the cohesion of \cref{eq:cohesion} and $\mathcal{G}^{v}_j=\Pi(S^{v}_j)$ in the
whitened space. The encoder of a round replaces the
current best only if it improves $\mathcal{H}$ by more than $\delta_{\mathcal{H}}$. The initial
encoder is itself a candidate, so adaptation can never be forced to degrade it. The loop stops
after $R_{\mathrm{p}}$ rounds without improvement. The number of anchors follows the same principle as
$K$: the loop runs for each $N_a\in\{16,24,32,48\}$, and we keep the run with the highest mean silhouette
(\cref{eq:silhouette}) of the average-linkage partition of $\{e_i\}$ on the validation sets.

\subsection{Source Partitioning}
\label{sec:partition}
Given a set $S$ of images to attribute, the final embeddings $\tilde{e}_i$ (adapted encoder and
whitening) are clustered by average-linkage agglomerative clustering on the cosine distance
\begin{equation}
  \mathrm{dist}(i,j)=1-\tilde{e}_i^{\top}\tilde{e}_j .
  \label{eq:distance}
\end{equation}
Cutting the dendrogram at $K'$ clusters gives a candidate partition $\mathcal{C}_{K'}$. For image $i$
in cluster $C(i)$ of $\mathcal{C}_{K'}$, let $a(i)$ be its mean distance (\cref{eq:distance}) to the
other members of $C(i)$ and $b(i)$ its smallest mean distance to another cluster. The
silhouette~\cite{rousseeuw1987silhouettes} of the candidate is
\begin{equation}
  \mathrm{sil}(K')=\frac{1}{|S|}\sum_{i\in S}\frac{b(i)-a(i)}{\max\{a(i),b(i)\}}.
  \label{eq:silhouette}
\end{equation}
On sets of 100 images the silhouette alone overestimates $K$ (\cref{tab:ablation}) and fails when
generators share one resolution, so we anchor $K$ to a catalog
of source identities.

\paragraph{Identity catalog.}
On the pool embeddings $\{e_i\}$, before whitening, the density of image $i$ is its mean cosine
similarity to its $k_d$ nearest neighbors, $\varrho_i$. Visiting images by decreasing density, each
image joins the group of its densest mutual neighbor already visited, or opens a new group; when
two groups meet at image $i$, they merge if the lower of their two peak densities exceeds $\varrho_i$
by less than $\eta\,\mathrm{std}(\varrho)$, as in persistence-based clustering~\cite{chazal2013tomato}.
Groups smaller than $\max(10,0.01\,|\mathcal{X}|)$ are dropped, and a group becomes an identity only
if it reappears, with Jaccard overlap of at least 0.5, in at least half of $B_s$ random halves of the
pool. An identity is represented by all its member images, so it can take any shape.

\paragraph{Number of sources.}
In a set $S$, each image votes for the identities of its $v$ nearest catalog members, weighted by
cosine similarity, and an identity is active if it wins at least $\max(3,0.05\,|S|)$ images. With
$A$ active identities,
\begin{equation}
  \Khat=\arg\max_{A+1\le K'\le A+4}\ \mathrm{sil}(K'),\qquad \hat{\mathcal{C}}=\mathcal{C}_{\Khat}.
  \label{eq:khat}
\end{equation}
The identity catalog tends to merge closely related generators, so the window starts above $A$; within it,
the silhouette decides. The window width is an implementation choice fixed on validation; deriving it
from the identities alone is left to future work.

\paragraph{Use of labels.}
Every training objective (\cref{eq:predloss,eq:vicreg,eq:pair,eq:adapt}), every checkpoint selection
(validation $\mathcal{L}_{\mathrm{pred}}$, validation $\mathcal{L}_{\mathrm{VIC}}$, and
\cref{eq:criterion}), the silhouette for $N_a$, the identity catalog, and every stopping rule and sampling step is computed without source labels.
We verified this by auditing each stage of the implementation. Hyper-parameters were chosen on the
validation sets and kept fixed for all datasets and experiments.

\paragraph{Implementation.}
$P_\theta$: embedding size 64, 4 heads, AdamW ($10^{-3}$), up to 100 epochs with early stopping.
$f_\phi$: AdamW ($10^{-3}$), batch 64. Adaptation: $J=120$ sets of $n=100$ images per round,
$R=25$, 20 epochs per round, $n_{\min}=12$, $\delta=0.02$, $T=0.1$, $\lambda=0.05$, $J_v=20$,
$\delta_{\mathcal{H}}=0.002$, $R_{\mathrm{p}}=8$, $J_w=60$. Pre-training: $k_q=3$, 8 rounds of 20 epochs,
batches of 128 pairs, AdamW ($3\cdot10^{-4}$). Partitioning: $k_d=20$, $\eta=0.5$, $B_s=10$, $v=10$, $K'\le15$.

%% file: sec/4_experiments.tex
\section{Experiments}
\label{sec:experiments}

We evaluate \ours on two datasets against label-free and semi-supervised methods.

\subsection{Experimental Setup}
\label{sec:setup}

\paragraph{Datasets.}
We collect face images from 20 generators, 1000 per generator. To compare fairly with
open-world methods, which need labeled known generators, and to test generalization to generators
never seen during training, the 20 generators are split into two disjoint groups of ten, \dsone
and \dstwo: when one is evaluated, the other provides the labeled generators those methods
require. \dsone gathers recent commercial and open diffusion services (Adobe Firefly~\cite{firefly},
DALL-E~3~\cite{dalle3}, Flux.1~\cite{flux2024}, Flux.1.1~Pro~\cite{fluxpro}, Freepik~\cite{freepik},
Leonardo~AI~\cite{leonardoai}, Midjourney~\cite{midjourney}, SD~3.5~\cite{stable_diffusion_35_large},
SDXL~\cite{sdxl_turbo}, Starry~AI~\cite{starryai}), all rendering the
same portrait prompts; we use the split released with it~\citep{bongini2025wild}, by prompt (50/20/30\% for
train/validation/test), so no content is shared between training and test. \dstwo mixes diffusion,
autoregressive and GAN generators (DALL-E~1~\cite{ramesh2021zeroshot}, DeepAI~\cite{deepai},
Hotpot~AI~\cite{hotpotai}, Sana-PAG~\cite{xie2024sana}, Stable Cascade~\cite{stablecascade}, SD
Attend-and-Excite~\cite{sd_ae}, StyleGAN~\cite{karras2019stylegan}, StyleGAN2~\cite{karras2020stylegan2},
StyleGAN3~\cite{karras2021stylegan3}, Hunyuan~\cite{li2024hunyuandit}) and is split 70/15/15\% per generator.
No labels are used, but all methods train on training images and select with label-free criteria on validation images; test images stay unseen until evaluation.

\paragraph{Protocol and metrics.}
We attribute 20 test sets of $n=100$ images drawn uniformly at random, so that every set contains
essentially all ten generators. No method is told the number of sources, except CDAL, which
requires it. We report
\begin{enumerate*}[label=(\roman*)]
  \item partition metrics: ARI~\cite{hubert1985ari}, NMI, pairwise F1 and $\Khat$;
  \item per-source metrics: the purity of clusters, the recovery of each generator (share of its
  images in a cluster where it is the majority), and the share of images in mixed clusters;
  \item representation metrics, independent of any clustering: pair AUC and leave-one-out 1-NN
  accuracy.
\end{enumerate*}
All values are averages over the 20 sets.

\paragraph{Compared methods.}
We compare \ours with the most recent methods that group or attribute synthetic images, published in
2025--2026, and with general-purpose foundation features: (i)~FSD~\cite{fsd}, the most recent
forensic descriptor supporting clustering by source, computed with the authors' code and
preprocessing (1024~px); (ii)~DROWCULA~\cite{drowcula}, which clusters
foundation features, with the authors' recipe (DINOv2-g~\cite{oquab2024dinov2} features, UMAP,
$k$-means and $K$ by the silhouette); (iii)~CDAL~\cite{cdal} and (iv)~CAL~\cite{cal}, the latest
open-world attribution methods, run with the code and hyper-parameters released by the authors. As
open-world methods, they need labeled known classes, so we give them the ten \emph{labeled}
generators of the other dataset and the target dataset unlabeled; CDAL is given $K$, CAL estimates
it. This privileged information is rarely available to an analyst: we report them as a reference. Finally, (v)~the image
embeddings of DINOv2 ViT-B/14 and ViT-g/14~\cite{oquab2024dinov2} and CLIP
ViT-L/14~\cite{radford2021clip}, clustered by average linkage with $K$ chosen by the silhouette, serve
as baselines.

\subsection{Main Results}
\label{sec:main}

\begin{table*}[t]
\centering
\caption{\textbf{Results on \dsone.} Averages over 20 test sets of 100 images (true $K=10$). Venue:
where each compared method was published. Gray: trained with ten labeled generators ($\dagger$).
Bold: best; underlined: second best. The last two columns measure the representation alone. \ours:
mean of three seeds (std of ARI 0.029 on \dsone, 0.015 on \dstwo); compared methods: one run with the
authors' code and settings.}
\label{tab:closed}
\setlength{\tabcolsep}{4pt}
\resizebox{\textwidth}{!}{%
\begin{tabular}{l ccc ccccccc cc}
\toprule
 & & & & \multicolumn{7}{c}{Partition} & \multicolumn{2}{c}{Representation} \\
\cmidrule(lr){5-11}\cmidrule(lr){12-13}
Method & Venue & Labels & $K$ given & ARI$\uparrow$ & NMI$\uparrow$ & Pair F1$\uparrow$ & Purity$\uparrow$ & Recovery$\uparrow$ & Mixed$\downarrow$ & $\Khat$ & Pair AUC$\uparrow$ & 1-NN$\uparrow$ \\
\midrule
DINOv2-B~\cite{oquab2024dinov2} & -- & -- & -- & 0.036 & 0.236 & 0.152 & 0.400 & 0.060 & 0.895 & \underline{9.15} & 0.636 & 0.614 \\
DINOv2-g~\cite{oquab2024dinov2} & -- & -- & -- & 0.021 & 0.193 & 0.143 & 0.341 & 0.028 & 0.945 & 7.85 & 0.609 & 0.545 \\
CLIP ViT-L~\cite{radford2021clip} & -- & -- & -- & 0.109 & 0.369 & 0.202 & 0.493 & 0.203 & 0.717 & 11.85 & 0.730 & 0.700 \\
DROWCULA~\cite{drowcula} & AJCAI'25 & -- & -- & 0.014 & 0.119 & 0.154 & 0.261 & 0.011 & 0.975 & 4.55 & 0.609 & 0.545 \\
FSD~\cite{fsd} & CVPR'25 & -- & -- & 0.258 & 0.525 & 0.375 & 0.368 & 0.009 & 0.979 & 2.80 & \textbf{0.913} & \textbf{0.904} \\
\midrule
\color{gray}CDAL$^\dagger$~\cite{cdal} & \color{gray}ICCV'25 & \color{gray}10 gen. & \color{gray}\checkmark & \underline{0.327} & \underline{0.555} & \underline{0.393} & \underline{0.648} & \underline{0.412} & \underline{0.404} & \color{gray}12.35 & \color{gray}0.832 & \color{gray}0.697 \\
\color{gray}CAL$^\dagger$~\cite{cal} & \color{gray}AAAI'26 & \color{gray}10 gen. & \color{gray}-- & \color{gray}0.288 & \color{gray}0.537 & \color{gray}0.383 & \color{gray}0.641 & \color{gray}0.250 & \color{gray}0.649 & \color{gray}8.85 & \color{gray}0.838 & \color{gray}0.699 \\
\midrule
\textbf{\ours} & -- & -- & -- & \textbf{0.535} & \textbf{0.721} & \textbf{0.588} & \textbf{0.726} & \textbf{0.586} & \textbf{0.185} & \textbf{9.35} & \underline{0.909} & \underline{0.872} \\
\bottomrule
\end{tabular}}
\end{table*}

\begin{table*}[t]
\centering
\caption{\textbf{Results on \dstwo.} Same protocol and notation as \cref{tab:closed}.}
\label{tab:open}
\setlength{\tabcolsep}{4pt}
\resizebox{\textwidth}{!}{%
\begin{tabular}{l ccc ccccccc cc}
\toprule
 & & & & \multicolumn{7}{c}{Partition} & \multicolumn{2}{c}{Representation} \\
\cmidrule(lr){5-11}\cmidrule(lr){12-13}
Method & Venue & Labels & $K$ given & ARI$\uparrow$ & NMI$\uparrow$ & Pair F1$\uparrow$ & Purity$\uparrow$ & Recovery$\uparrow$ & Mixed$\downarrow$ & $\Khat$ & Pair AUC$\uparrow$ & 1-NN$\uparrow$ \\
\midrule
DINOv2-B~\cite{oquab2024dinov2} & -- & -- & -- & 0.215 & 0.463 & 0.309 & 0.534 & 0.169 & 0.714 & 9.75 & 0.769 & 0.681 \\
DINOv2-g~\cite{oquab2024dinov2} & -- & -- & -- & 0.173 & 0.438 & 0.263 & 0.528 & 0.170 & 0.732 & 11.40 & 0.733 & 0.621 \\
CLIP ViT-L~\cite{radford2021clip} & -- & -- & -- & 0.186 & 0.419 & 0.306 & 0.471 & 0.071 & 0.881 & 4.70 & 0.840 & 0.785 \\
DROWCULA~\cite{drowcula} & AJCAI'25 & -- & -- & 0.139 & 0.339 & 0.253 & 0.420 & 0.070 & 0.876 & 6.10 & 0.733 & 0.621 \\
FSD~\cite{fsd} & CVPR'25 & -- & -- & 0.304 & 0.584 & 0.412 & 0.383 & 0.030 & 0.925 & 2.90 & 0.927 & \textbf{0.934} \\
\midrule
\color{gray}CDAL$^\dagger$~\cite{cdal} & \color{gray}ICCV'25 & \color{gray}10 gen. & \color{gray}\checkmark & \textbf{0.648} & \underline{0.788} & \textbf{0.682} & \underline{0.788} & \textbf{0.729} & \textbf{0.079} & \color{gray}12.10 & \underline{0.935} & \color{gray}0.826 \\
\color{gray}CAL$^\dagger$~\cite{cal} & \color{gray}AAAI'26 & \color{gray}10 gen. & \color{gray}-- & \color{gray}0.427 & \underline{0.668} & \color{gray}0.496 & \color{gray}0.670 & \color{gray}0.369 & \color{gray}0.369 & \textbf{9.85} & \color{gray}0.920 & \color{gray}0.790 \\
\midrule
\textbf{\ours} & -- & -- & -- & \underline{0.590} & \textbf{0.794} & \underline{0.638} & \textbf{0.807} & \underline{0.602} & \underline{0.278} & \underline{9.73} & \textbf{0.946} & \underline{0.898} \\
\bottomrule
\end{tabular}}
\end{table*}

\begin{figure*}[t]
  \centering
  \includegraphics[width=0.85\textwidth]{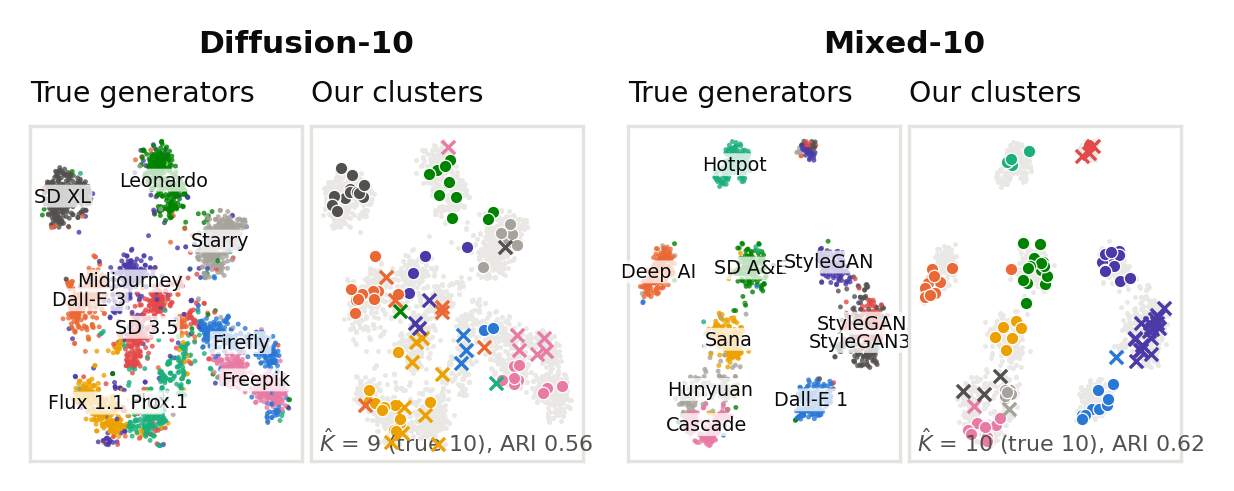}
  \caption{\textbf{Clusters found by \ours.} Left: t-SNE of all test embeddings, colored by true
  generator. Right, on the same map: one of the 20 test sets (100 images, closest to the mean ARI and
  $\Khat$), colored by the generator matched to each cluster; $\times$: image of another generator.}
  \label{fig:clusters}
\end{figure*}

\cref{tab:closed,tab:open} report the main comparison. Among label-free methods, \ours is best on
every partition and per-source metric on both datasets. It roughly doubles the ARI of FSD (0.535
vs.\ 0.258 and 0.590 vs.\ 0.304) and recovers about 60\% of each generator's images in its own
cluster, against at most 20\% for the other label-free methods. FSD has the highest 1-NN accuracy, but its descriptor
does not form clusters: FSD collapses to 2--3 groups. The open-world methods, despite ten labeled generators, fall behind \ours
on \dsone, whose new generators are closely related diffusion services. On \dstwo, CDAL, which is
also given $K$, attains a higher ARI. \ours has a higher NMI, purity and pair AUC, and
outperforms CAL by 0.16 ARI.
\cref{fig:clusters} shows a typical test set: without knowing $K$, \ours finds nine and ten
clusters, and most generators form their own; errors concentrate on the related diffusion services of
\dsone and on the three StyleGANs, which merge into one cluster.

\subsection{How Many Generators?}
\label{sec:k}

\begin{figure}[t]
  \centering
  \includegraphics[width=0.55\linewidth]{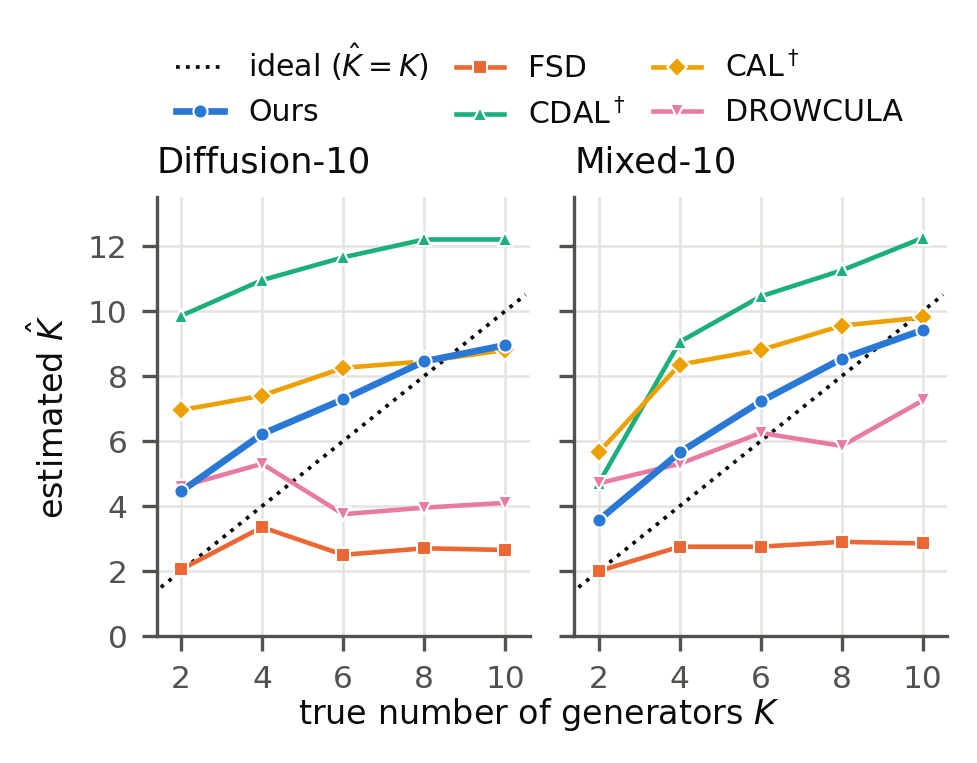}
  \caption{\textbf{$\Khat$ vs.\ the true number of generators} in sets of 100 images containing
  exactly $K$ generators. \ours has the smallest mean deviation from the diagonal, but
  overestimates $K$ when few generators are present.}
  \label{fig:kgen}
\end{figure}

\ours estimates $K$ to within about one generator on average ($\Khat=9.35$ and $9.73$). FSD and DROWCULA
collapse to 2--6 clusters, CAL underestimates $K$ on \dsone, and CDAL overestimates it although it is given
$K$. We then vary the number of generators per set from 2 to 10, composing the sets with labels
(\cref{fig:kgen}). \ours grows with the true value and has the smallest mean deviation from it
(1.5 and 1.1 generators), but it overestimates $K$ when few generators are present ($\Khat=4.5$ and
$3.6$ for $K=2$): the window of \cref{eq:khat} starts above the number of active identities, and the
extra clusters split a generator. FSD and DROWCULA stay nearly constant; CAL and CDAL grow with $K$
but stay farther from the diagonal.

\subsection{Representation and Interpretability}
\label{sec:interpretability}

Before any clustering, \ours also separates generators better on average than both
semi-supervised methods (1-NN 0.872 vs.\ 0.697 and 0.699 on \dsone; 0.898 vs.\ 0.826 and 0.790 on \dstwo). The largest gaps occur on similar commercial services (Starry~AI, DALL-E~3, SD~3.5) and on the StyleGAN family. Per-generator
cluster F1 confirms it: \ours is best on 8 of 10 generators of
\dsone, CDAL on 7 of 10 of \dstwo.

Since all generators of \dsone share the same prompts, we also checked what drives similarity:
images with the same prompt from different generators are unrelated in our space, while images from
the same generator with different prompts are close, the opposite of DINOv2.

\subsection{Robustness to Post-Processing}
\label{sec:robustness}

\begin{table}[t]
\centering
\caption{\textbf{ARI under chained post-processing} (1--3 random operations among eleven, including crop,
blur, resize, JPEG, WebP and JPEG-AI). Inference only: every model comes from clean
data. Clean: same image subset as each level. Bold: best; underlined: second best.}
\label{tab:robustness}
\setlength{\tabcolsep}{3pt}
\small
\begin{tabular}{l c cccc cccc}
\toprule
 & & \multicolumn{4}{c}{\dsone} & \multicolumn{4}{c}{\dstwo} \\
\cmidrule(lr){3-6}\cmidrule(lr){7-10}
Method & Venue & clean & 1 & 2 & 3 & clean & 1 & 2 & 3 \\
\midrule
DINOv2-B~\citep{oquab2024dinov2} & -- & 0.037 & 0.035 & 0.026 & 0.031 & 0.256 & 0.236 & 0.214 & 0.196 \\
DINOv2-g~\citep{oquab2024dinov2} & -- & 0.019 & 0.021 & 0.025 & 0.028 & 0.202 & 0.192 & 0.179 & 0.180 \\
CLIP ViT-L~\citep{radford2021clip} & -- & 0.142 & 0.081 & 0.061 & 0.051 & 0.197 & 0.182 & 0.180 & 0.190 \\
DROWCULA~\citep{drowcula} & AJCAI'25 & 0.010 & 0.009 & 0.007 & 0.016 & 0.168 & 0.157 & 0.165 & 0.141 \\
FSD~\citep{fsd} & CVPR'25 & 0.240 & 0.099 & 0.041 & 0.013 & 0.334 & 0.169 & 0.033 & 0.016 \\
\midrule
\color{gray}CDAL~\citep{cdal}$^\dagger$ & \color{gray}ICCV'25 & \underline{0.336} & \color{gray}0.187 & \color{gray}0.096 & \color{gray}0.059 & \textbf{0.652} & \textbf{0.511} & \textbf{0.411} & \textbf{0.297} \\
\color{gray}CAL~\citep{cal}$^\dagger$ & \color{gray}AAAI'26 & \color{gray}0.299 & \underline{0.224} & \underline{0.168} & \textbf{0.122} & \color{gray}0.420 & \color{gray}0.361 & \underline{0.322} & \underline{0.289} \\
\midrule
\textbf{\ours} & -- & \textbf{0.548} & \textbf{0.313} & \textbf{0.176} & \underline{0.080} & \underline{0.591} & \underline{0.443} & \underline{0.322} & 0.228 \\
\bottomrule
\end{tabular}
\end{table}

\begin{table}[t]
\centering
\caption{\textbf{Ablations} (test sets; ARI clean and at post-processing level 3, $\Khat$ with true $K=10$;
three seeds, one for the VICReg encoder). Bottom: trace input, measured with the previous adaptation.}
\label{tab:ablation}
\setlength{\tabcolsep}{3pt}
\small
\begin{tabular}{l ccc ccc}
\toprule
 & \multicolumn{3}{c}{\dsone} & \multicolumn{3}{c}{\dstwo} \\
\cmidrule(lr){2-4}\cmidrule(lr){5-7}
Variant & ARI & step 3 & $\Khat$ & ARI & step 3 & $\Khat$ \\
\midrule
\textbf{\ours (full)}              & \underline{0.535} & 0.080 & \underline{9.35} & \underline{0.590} & 0.228 & \textbf{9.73} \\
$K$ by the silhouette alone        & \textbf{0.550} & -- & 10.75 & \textbf{0.601} & -- & 10.85 \\
no mutual-neighbor stage          & 0.513 & -- & \textbf{9.42} & 0.553 & -- & \underline{9.42} \\
no self-training loop              & 0.421 & -- & 7.55 & 0.470 & -- & 7.88 \\
no adaptation (VICReg encoder)        & 0.256 & -- & 4.90 & 0.448 & -- & 7.75 \\
\midrule
residual $r_i$ (previous)          & 0.491 & 0.059 & 10.20 & 0.586 & 0.217 & 10.10 \\
latent $z_i$ instead of $r_i$      & 0.430 & 0.027 & 9.95 & 0.557 & 0.107 & 8.90 \\
\bottomrule
\end{tabular}
\end{table}

\cref{tab:robustness} chains 1 to 3 operations, drawn at random for each test image from eleven
common edits such as compression, resizing, cropping and blurring. \ours is the most robust label-free method at every level, and on \dsone it also leads
both open-world methods up to two operations. FSD drops close to zero
after two operations. DROWCULA and the foundation features barely change, but only because they
start low. With three operations, CAL on \dsone and CDAL and CAL on \dstwo
retain more accuracy; retrained with two labeled generators, both fall well below \ours; their extra robustness comes from supervision no label-free method can use.

\subsection{Ablations}
\label{sec:ablation}
\textbf{Residuals vs.\ latents} (\cref{tab:ablation}, bottom). Feeding the raw latent $z_i$ to the
encoder lowers the clean ARI and halves the accuracy under heavy post-processing. The masked
residual is what makes the trace stable.
\textbf{Adaptation.} Both stages are needed: without the mutual-neighbor stage ARI drops by 0.02
and 0.04, without the self-training loop by 0.11 and 0.12.
\textbf{Number of sources.} Here the silhouette alone is as good (ARI $+0.01$); the identity catalog matters
when generators share one resolution, raising ARI from 0.38 to 0.47 and from 0.46 to 0.63.

%% file: sec/5_conclusion.tex
\section{Discussion and Limitations}
\label{sec:discussion}

Open-world methods trained on ten
labeled generators reach a higher ARI on \dstwo and degrade less under heavy post-processing. With two
labeled generators both advantages vanish, and across set
compositions they estimate $K$ less reliably than \ours (\cref{fig:kgen}). Their lead relies on information a real case lacks. Without labels, \ours still separates generators better than both (1-NN in
\cref{tab:closed,tab:open}).

Heavy post-processing is the main limit of \ours: after three
chained operations ARI falls from 0.55 to 0.08 on \dsone, a larger drop than for CAL even with
five labeled generators. Compact families are the second: the
three StyleGANs are the hardest generators for every method. Third, the window of \cref{eq:khat}
overestimates $K$ when a set holds only a few generators (\cref{fig:kgen}). Finally, the adaptation
is transductive: it needs an unlabeled pool from the same generators as the test collection. When
three generators are left out of the pool, about a quarter of their images still form their own
cluster, and the rest join known generators or mixed clusters. Faces are our scope by design: on GenImage, whose generators render a thousand object
classes, the clusters do not follow the generators (ARI 0.08), and
separating the trace from varied content needs a different design.

\section{Conclusion}
\label{sec:conclusion}

We introduced label-free open-world source attribution: partitioning AI-generated images by
generator without labels and without knowing how many there are. \ours addresses it with a masked-prediction residual in a frozen latent
space and a transductive label-free adaptation that makes this trace, not content, drive
similarity. On two datasets of ten
generators, \ours roughly doubles the ARI of the strongest label-free method, estimates their number
to within about one, also with a single shared resolution, and outperforms CAL on both datasets and
CDAL on one, despite their labels. It is also the most robust label-free method under post-processing.